\documentclass[twocolumn]{svjour3}         
\smartqed  
\usepackage{graphicx}

\usepackage{titlesec} 
\titleformat{\paragraph}[runin]
{\bfseries\scshape}{\theparagraph}{1em}{}
\titlespacing{\paragraph}{0em}{1ex}{.5em} 
\usepackage{todonotes}
\usepackage{booktabs}
\usepackage{tabularx}
\usepackage{multirow}
\usepackage{multicol}
\usepackage{bm}
\usepackage{url}
\usepackage{balance}
\usepackage{enumitem}
\usepackage{color}
\usepackage{bbding}
\usepackage{nccmath}
\usepackage{amsmath,amssymb}
\usepackage{graphicx}
\usepackage{cite}
\begin{document}

\title{View Birdification in the Crowd
}
\subtitle{Ground-Plane Localization from Perceived Movements}


\author{Mai Nishimura \and
        Shohei Nobuhara \and
        Ko Nishino 
}


\institute{Mai Nishimura \at
             Graduate School of Informatics, Kyoto University, Kyoto, Japan \\
             OMRON SINIC X, Tokyo, Japan.\\
              \email{mai.nishimura@sinicx.com}
           \and
           Shohei Nobuhara \at
           Graduate School of Informatics, Kyoto University, Kyoto, Japan \\
              \email{nob@i.kyoto-u.ac.jp}
           \and
           Ko Nishino \at
           Graduate School of Informatics, Kyoto University, Kyoto, Japan \\
              \email{kon@i.kyoto-u.ac.jp}
}

\date{Received: date / Accepted: date}

\maketitle


\def\eg{\emph{e.g.}}
\def\etal{\emph{et al}}

\newcommand{\EQREF}{Eq.~\eqref}
\newcommand{\EQSREF}{Eqs.~\eqref}
\newcommand{\FIGREF}{Fig.~\ref}
\def\proposed{VB} 
\def\fixcolor{black}
\def\hstate{\bm {\tilde s}}
\def\rstate{\bm s}
\def\jstate{\bm s^{jn}}
\def\jpolicy{\overrightarrow{\pi}}
\def\vpref{v_{\text {pref}}}
\def\vector#1{\mbox{\boldmath $#1$}}
\def\sup#1{^{(\rm #1)}}
\def\sub#1{_{\rm #1}}
\def\supi#1{^{(#1)}}
\def\vct#1{\mbox{\boldmath $#1$}}
\def\eg{{\it e.g.}}
\def\cf{{\it c.f.}}
\def\ie{{\it i.e.}}
\def\etal{{\it et al. }}
\def\etc{{\it etc}}
\newcommand{\argmax}{\mathop{\rm argmax}\limits}
\newcommand{\argmin}{\mathop{\rm argmin}\limits}

\def\Rerr{\Delta \bm r}
\def\Terr{\Delta \bm t}
\def\Xerr{\Delta \bm x}
\def\XerrRel{\Delta \bm {\tilde x}}
\def\Xgt{\dot{\bm x}}
\def\Rgt{\dot{R}}
\def\Tgt{\dot{\bm t}}
\def\arraystretchlen{1.0}

\def\cam{c}
\def\image{\mathcal I}
\def\traj{\mathcal X}
\def\btraj{\mathcal {\bm X}}
\def\keypoints{\mathcal P}
\def\states{\mathcal S}
\def\bstates{\mathcal {\bm S}}
\def\state{\bm s}
\def\ped{\bm x}
\def\pedi{\bm p} 
\def\obs{\bm z}

\def\Fi{\bm F_r}
\def\Fp{\bm F_p}
\def\vpref{\bm w}
\def\ENERGY{{\mathcal E}}

\def\DIFF#1{\textcolor{black}{#1}}
\def\DIFFCR#1{\textcolor{black}{#1}}

\begin{abstract}
We introduce \emph{view birdification}, the problem of recovering ground-plane movements of people in a crowd from an ego-centric video captured from an observer (\eg, a person or a vehicle) also moving in the crowd. Recovered ground-plane movements would provide a sound basis for situational understanding and benefit downstream applications in computer vision and robotics. In this paper, we formulate view birdification as a geometric trajectory reconstruction problem and derive a cascaded optimization method from a Bayesian perspective. The method first estimates the observer's movement and then localizes surrounding pedestrians for each frame while taking into account the local interactions between them. We introduce three datasets by leveraging synthetic and real trajectories of people in crowds and evaluate the effectiveness of our method. The results demonstrate the accuracy of our method and set the ground for further studies of view birdification as an important but challenging visual understanding problem.
\end{abstract}

\section{Introduction}
We as human beings are capable of mentally visualizing our surroundings from a third-person view. Imagine walking down a street alongside other pedestrians. Your mental model of the surrounding movements of people is not a purely two-dimensional one, but rather in 3D, albeit imperfect, in which you can virtually fly around. It lets you anticipate potential collisions so that you can avoid them or guess the goal of another person so that you can follow. Some people have exceptionally high capabilities in forming such a virtual view (\eg, a professional soccer player), but nonetheless, we all rely on this 3D spatial sense to complement our ego-centric view in our daily lives.
\begin{figure}
  \includegraphics[width=\linewidth]{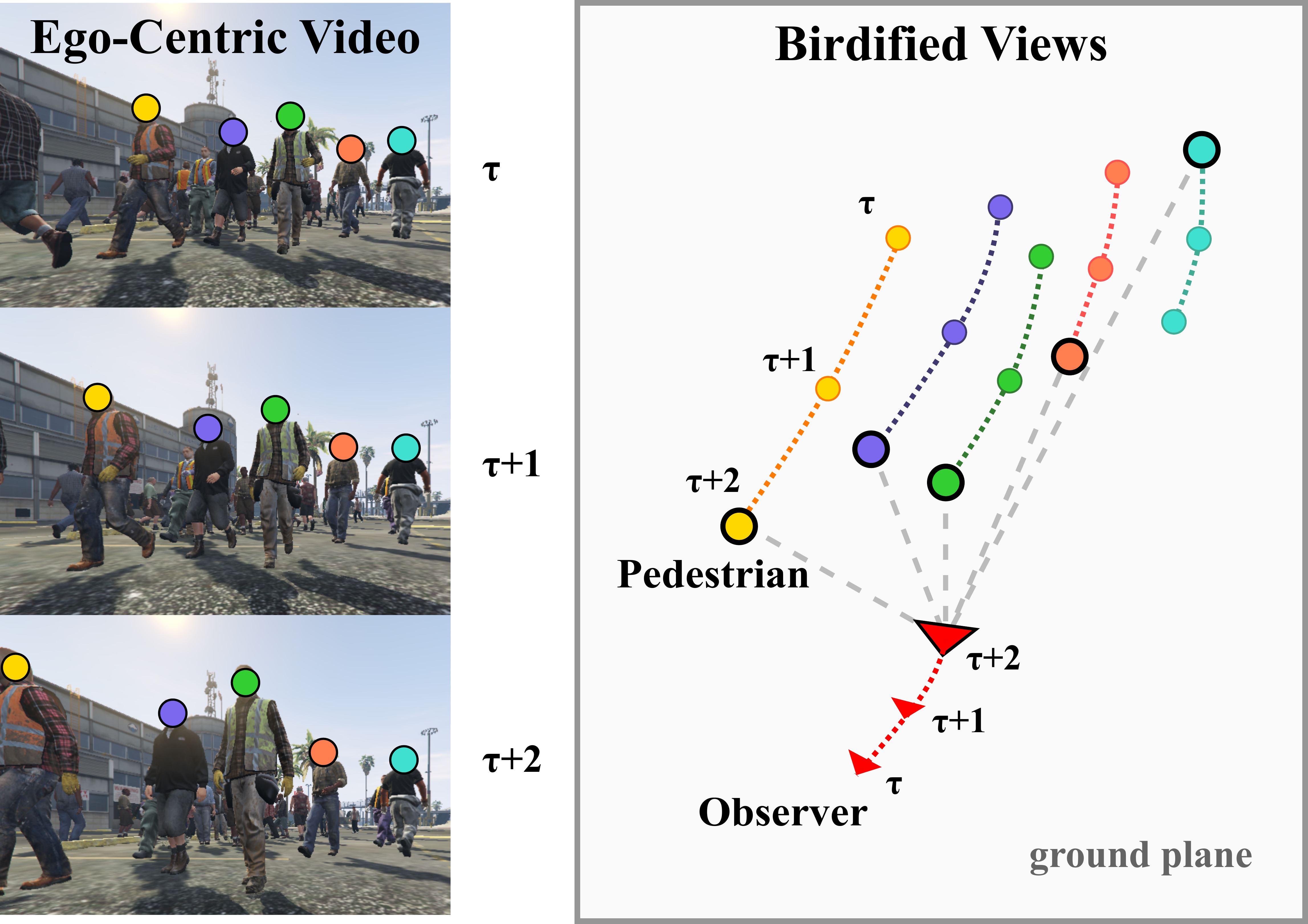}
\caption{We introduce view birdification of a crowd, the task of estimating the movements of surrounding people on the ground plane (right) from a single dynamic ego-centric image sequence (left), and derive a stratified optimization method based on the geometric relations of pedestrians' projections and interactions.}
\label{fig:teaser}
\end{figure}
Endowing such global 3D spatial perception with computers, however, remains elusive. Despite the significant progress in computational 3D and motion perception, including stereo, structure from motion, and optical flow estimation, a bottom-up approach of first reconstructing the 3D geometry and motion and then changing the viewpoint would be brittle. Its success would inherently hinge on the accuracy of each step which is prone to fundamental ambiguities between them. Can we bypass these and directly obtain a virtual perspective of the surroundings? More specifically, can we recover the dynamically changing global layout of people moving around ourselves solely from images captured from our vantage point while we also move around?

In this paper, we introduce \emph{view birdification} in a crowd~\cite{nishimura2021view}, the problem of computing a bird-eye's view of the movements of surrounding people from a single ego-centric view of a moving person (see \FIGREF{fig:teaser}) and derive a geometric solution to it. View birdification differs from recent works on bird-eye view rendering where the goal is to render a bird's eye view image from a given ego-centric image, \ie, view transformation such that the scene appearance is imaged fronto-parallel to the ground. We are, in contrast, interested in deciphering and laying out the movements of people on the ground plane from a temporal sequence of a dynamic ego-centric view. In other words, from a personal perspective of a crowd, we would like to see how everybody is moving (not how they look) as seen from the top (\ie, a bird hovering over the crowd).

View birdification of a crowd would have a wide range of applications. It will let us analyze the global and local interactions of people from a holistic perspective both in space and time, which would benefit areas such as navigation~\cite{nishimura2020iros,anvarios2020}, movement prediction~\cite{ivanovic2019trajectron,gupta2018social,alahi2016social}, and surveillance~\cite{kratz2009anomaly}. It can also offer a crucial visual perception for self-driving cars to gauge surrounding activities.

Unlike bird-eye view rendering which can be formulated as an image to image transformation (\eg, with a deep neural network), view birdification does not concern the appearance of the scene captured in the ego-centric view. From observations of dynamically moving objects, our method localizes the moving camera and simultaneously maps the dynamic objects on the ground plane. This is reminiscent of SLAM but with the fundamental difference that everything is dynamic. The dynamic objects (\ie, people) also do not embody any low-dimensional structure as often assumed in non-rigid structure from motion.

Our method is based on two key insights. First, the movements of dynamic keypoints (\eg, head-points of pedestrians) are not arbitrary, but exhibit coordinated motion that can be expressed with crowd flow models~\cite{helbing1995social,pellegrini2009you}. That is, the interaction of pedestrians' movements in a crowd can be locally described with analytic or data-driven models. Second, the scale and difference of human heights are proportional to estimated geometric depth~\cite{luo2020whether}. In other words, the positions of pedestrians on the ground plane can be constrained along the lines that pass through a center of projection. These insights lend us a natural formulation of view birdification as a geometric reconstruction problem.

We formulate view birdification as a geometric reconstruction problem and derive a solution based on stratified optimization. Our stratified optimization consists of the observer's camera ego-motion estimation with pedestrian movement interactions as pairwise constraints and pedestrian localization given the ego-motion estimate and height priors on the pedestrians. We first solve this camera ego-motion estimation by gradient descent and then localize each pedestrian given the observer's camera position as a combinational optimization problem with pairwise interaction constraints.

We experimentally validate our method on both synthetic and real trajectories extracted from publicly available crowd datasets.
We create a photorealistic crowd dataset that simulates real camera projection with a limited field of view and occluded pedestrian observations while moving in the crowd. These datasets allow us to quantitatively evaluate our method systematically. Experimental results demonstrate the effectiveness of our approach for view birdification in crowds of various densities. The results on the photorealistic crowd dataset show the end-to-end effectiveness of our method, from person detection to localization on the ground plane, demonstrating its performance in real-world use. 
We also test our method on real-robot dataset captured in crowds. The results show that our method can work both for body-worn cameras and mobile robot platforms.
We believe these results have strong implications in computer vision and robotics as they establish view birdification as a foundation for downstream visual understanding applications including crowd behavior analysis and robot navigation.

\section{Related Work}

To our knowledge, our work is the first to formulate view birdification. View birdification is related to a number of computer vision and robotics problems whose relevant works we review in this section.

\paragraph{Bird's Eye View Transformation}
Conceptually, view birdification may appear similar to bird's eye view (BEV) synthesis. These two are fundamentally different in three critical ways. First, view birdification concerns the movements not the appearance BEV synthesis
~\cite{zhou2016view,regmi2018cross,zhu2018generative,tang2019multichannel,tang2019local}
or cross-view association~\cite{ardeshir2016ego2top,ardeshir2016egotransfer,soran2014action}. Second, unlike most BEV methods~\cite{lin2012vision,taneja2010modeling,mustafa2015general}, view birdification should not rely on ground plane keypoints, multi-view images, or paired images between the views as they are usually not available in crowded scenes. Also note that, in crowded scenes, the ground plane and footsteps cannot be clearly extracted, which makes simple homography-based approaches impossible. Third, view birdification aims to localize all agents in a single coordinate frame across time, unlike BEV which is relative to the observer's location at each time instance~\cite{mani2020monolayout,bertoni2019monoloco,zhang2021body}. As such, BEV synthesis methods are not directly applicable to view birdification. These conceptual differences allow the birdification to better understand the dynamics of the surroundings including ego-motion in a challenging crowded environment.

\paragraph{Dynamic SLAM}
View birdification can be considered as a dynamic SLAM problem in which all points, not just the observer but also the scene itself, are dynamic. Typical approaches to dynamic SLAM explicitly track and filter dynamic objects~\cite{yu2018ds,bescos2018dynaslam} or implicitly minimize outliers caused by the dynamic objects~\cite{hahnel2002map,hahnel2003map,lv2019taking}. In contrast to these approaches that sift out static keypoints from dynamic ones, methods that leverage both static and dynamic keypoints by, for instance, constructing a Bayesian factor graph~\cite{li2018stereo,henein2020dynamic,huang2020clustervo} have also been introduced. The success of most of these approaches, however, depends on static keypoints which are hard to find and track in cluttered dynamic scenes such as in a dense crowd. In view birdification, we require no static keypoints and can reconstruct both ego-motion and surrounding dynamics only from the observed motions in the ego-view.

\paragraph{Crowd Modeling}
Modeling human behavior in crowds is essential for a wide range of applications including crowd simulation~\cite{lerner2007crowds}, trajectory forecasting~\cite{alahi2016social,ivanovic2019trajectron,gupta2018social}, and robotic navigation~\cite{nishimura2020iros,tai2018socially,anvarios2020}. Popular approaches include multi-agent interactions based on social force models~\cite{helbing1995social,mehran2009abnormal,anvarios2020}, reciprocal force models~\cite{van2011reciprocal}, and imitation learning~\cite{tai2018socially}.
Recently, data-driven approaches have achieved significant performance gains on public crowd datasets~\cite{alahi2016social,gupta2018social,ivanovic2019trajectron}. All these approaches, however, are only applicable to near top-down views. Forecasting the future location of people from first-person viewpoints has also been explored~\cite{yagi2018future,MCBB20}, but they are limited to localization in the image plane.
View birdification may provide a useful foundation for these crowd modeling tasks.

\paragraph{Non-rigid Structure from Motion}
Reconstruction of point trajectories is also studied in the literature on non-rigid structure from motion (NRSfM)~\cite{jensen2020benchmark,saputra2018visual}, in particular as multi-body~\cite{kumar2016multi} and trajectory-based~\cite{akhter2008nonrigid,park20123d} approaches. NRSfM exploits the inherent global dynamic structure embodied by the target surface and the camera motion. In contrast, our focus is pedestrian trajectories that interact locally in an on-the-fly manner and do not exhibit coherent global structures that we can leverage.

\section{Geometric View Birdification}
A typical scenario for view birdification is when a person with a body-worn camera is immersed in a crowd consisting of people heading towards their destinations while implicitly interacting with each other. Our goal is to deduce the global movements of people from the local observations in the ego-centric video captured by a single person. 

\subsection{Problem Setting}
As a general setup, we assume that $K$ people are walking on a ground plane and an observation camera is mounted on one of them. We set the $z$-axis of the world coordinate system to the normal of the ground plane ($x$-$y$ plane) and denote the on-ground location of the $k^{\text{th}}$ pedestrian as $\ped_k = \left [x_k, y_k \right ]^\top$. Let us denote the location of $0^{\text{th}}$ person in the crowd $\ped_0$ as the observer capturing the ego-centric video of pedestrians $k \in \{1, 2, \dots, K\}$ who are visible to the observer. The observation camera is located at $\left [ x_0,y_0,h_0\right ]^\top$, where the mounted height $h_0$ is constant across the frames. We assume that the viewing direction is parallel to the ground plane, \eg, the person has a camera mounted on the shoulder. The same assumption applies when the observer is a vehicle or a mobile robot. At each timestep $\tau$, the pedestrians are observed by a camera with the pose $\left [ R^\tau | \bm t^\tau \right ]$ consisting of rotation matrix $R$ and translation vector $\bm t$.
\DIFF{
Assuming that the viewing direction of the camera is stabilized and parallel to the ground plane, we can approximate the rotation angles about the $x$- and $y$-axis to be $0$ across the frames. That is, the camera pose to be estimated is represented by its 2D rotation about $z$-axis $R(\Delta \theta_z) \in SO(2)$ and 2D translation $\bm t = -R(\Delta \theta_z)\Delta \ped_0 \in \mathbb R^2$ on the $x$-$y$ plane.}

We assume that bounding boxes of the people captured in the ego-video are already extracted. For this, we can use an off-the-shelf multi-object tracker~\cite{xiu2018poseflow,wang2019towards} which provides the state of each pedestrian on the image plane $\bm s_k^\tau = \left [ u_k^\tau, v_k^\tau, l_k^\tau\right ]^\top$ which consists of the projections of center location and height, $\pedi_k^\tau = \left [u_k^\tau,v_k^\tau\right ]^\top$ and $l_k^\tau$, respectively. Note that our method is agnostic to the actual tracking algorithm. Pedestrian IDs $k \in \{1,2,\dots,K\}$ can also be assigned by the tracker. Given a sequence of pedestrian states $\states_k$ from the first visible frame $\tau_1$ to the last visible frame $\tau_2$, \ie\,, $\states_k^{\tau_1:\tau_2} = \{ \state^{\tau_1}_k, \state_k^{\tau_1+1}, \dots, \state_k^{\tau_2} \}$, our goal is to simultaneously reconstruct the $K$ trajectories of the surrounding pedestrians $\traj_k^{\tau_1:\tau_2}=\{ \ped_k^{\tau_1}, \ped_k^{\tau_1+1},\dots, \ped_k^{\tau_2} \}$ and that of the observation camera $\traj_0^{\tau_1:\tau_2}=\{ \ped_0^{\tau_1}, \ped_0^{\tau_1+1}, \dots, \ped_0^{\tau_2} \}$ with its viewing direction $\mathcal R^{\tau_1:\tau_2} = \{ R^{\tau_1}, R^{\tau_1+1},\dots,R^{\tau_2} \}$ on the ground plane.

\subsection{Geometric Observation Model}
We assume a regular perspective ego-centric view or a $360^\circ$ cylindrical projection view. The following derivation also applies to other linear projection models including generic quasi-central cameras for fish-eye lens~~\cite{brousseau2019calibration}. 
\paragraph{Perspective Projection Model}
\DIFF{
In the case of perspective projection with focal length $f$ and intrinsic matrix $A \in \mathbb R^{3\times 3}$, the distance of the pedestrian from the observer is proportional to the ratio of the pedestrian height $h_k$ and its projection $l_k$, $\ie, h_k/l_k$. 
Given the center projection of the} pedestrian in the image plane $\state_k = [u_k, v_k, l_k]$, the on-ground location estimate of the pedestrian relative to the camera $\bm z_k = [\tilde x_k, \tilde y_k, 0]^\top$ can be computed by inverse projection of the observed image coordinates,
\begin{equation}
     \begin{bmatrix}\tilde x_k & \frac{h_k}{2} & \tilde y_k \end{bmatrix}^\top = \frac{f h_k}{l_k} A^{-1}\begin{bmatrix}u_k & v_k & 1\end{bmatrix}^\top\,,
    \label{eq:inv-projection}
\end{equation}
where the intrinsics $A$ and focal length $f$ are known since the observation camera can be calibrated \DIFFCR{a priori. The relative coordinates $\bm z_k$ are thus scaled by the unknown pedestrian height parameter $h_k$.
} 

\paragraph{Cylindrical Projection Model}
Mobile platforms often use a $360^\circ$ panorama view for a full view of the surroundings, which are composed of synchronized RGB sensor images.
Given a stitched $360^\circ$ cylindrical image with image width $W$ and the observed pedestrian state $\state_k = [u_k,v_k,l_k]^\top$ in the image, the location angle $\phi$ [rad] for the pedestiran position $\pedi_k=[u_k, v_k]^\top$ on the cylinder circle becomes
\begin{equation}
  \phi = 2\pi \frac{u_k}{W} - \pi.
\end{equation}
The inverse projection depth from the center of the circle $\tilde {r}_k$ is proportional to the ratio of the pedestrian height $h_k$ and its projection $l_k$,
\begin{align}\label{eq:cylindrical-projection}
    \tilde {r}_k = \tilde y_k \sec (\phi) = f \frac{h_k}{l_k}.
\end{align}
The on-ground location estimates of the pedestrian can be recovered as $\bm z_k = [\tilde r_k \sin(\phi), \tilde r_k \cos(\phi), 0]^\top$.

\DIFF{The absolute position of the pedestrian $\ped_k = [x_k,y_k]^\top$ can be computed by the relative coordinates $\bm z_k = [\tilde x_k, \tilde y_k]^\top$, the camera position $\ped_0 = [x_0,y_0]^\top$, and the viewing direction $\theta_z$ about $z$-axis,
\begin{equation}
    \begin{bmatrix}
      x_k \\ y_k 
    \end{bmatrix}
    = R_z(\theta_z)^\top 
    \begin{bmatrix}
      \tilde x_k  \\ \tilde y_k 
    \end{bmatrix} +
    \begin{bmatrix}
      x_0 \\ y_0
    \end{bmatrix}\,.
\end{equation}}
In what follows, we assume the most general case, \ie, perspective projection. The optimization pipeline, however, can be applied to any type of linear projection model without major changes.

\section{A Cascaded Optimization for View Birdification}
In this section, we introduce a cascaded optimization approach to the geometric view birdification problem based on a Bayesian perspective.
We first describe the overall energy minimization framework and then derive energy functions to be optimized for the two typical models.

\subsection{A Bayesian Formulation}
When a frame is pre-processed to a set of states $\mathcal S_{1:K}^\tau = \{ \state_1^\tau, \state_2^\tau, \dots, \state_K^\tau \} \in \mathbb R^{3\times K}$ at time $\tau$, we obtain a set of on-ground position estimates relative to a camera $\mathcal {Z}_{1:K}^\tau = \{ \bm z^\tau_1, \bm z^{\tau}_2, \dots, \bm z^{\tau}_K\} \in \mathbb R^{2 \times K}$ corresponding to the states $\mathcal { S}_{1:K}^\tau$. Assuming that we have sequentially estimated on-ground positions up to time $\tau-1$, $\traj_{0:K}^{\tau_0:\tau-1} = \{\traj_0^{\tau_0:\tau-1}, \traj_1^{\tau_0:\tau-1}, \dots , \traj_K^{\tau_0:\tau-1}\} \in \mathbb R^{2\times (K+1) \times \Delta \tau}$ with a temporal window of $\Delta \tau$ and its initial timestamp $\tau_0 = \tau - \Delta \tau$,
the posterior probability of the on-ground positions $\mathcal {X}_{0:K}^\tau = \{ \ped_0^\tau, \ped_1^\tau, \dots, \ped_K^\tau \} \in \mathbb R^{2 \times (K+1)}$ at time $\tau$ can be factorized as
\begin{equation}
  \begin{split}
     &p(\traj^\tau_{0:K}|\mathcal {Z}^{\tau}_{1:K},\traj^{\tau_0:\tau-1}_{0:K})\\
     &\propto
     p(\traj^\tau_{0:K}|\traj^{\tau_0:\tau-1}_{0:K})
     p(\mathcal Z^\tau_{1:K}|\traj^\tau_{0:K},\traj_{0:K}^{\tau_0:\tau-1})\,.
    \label{eq:posterior}
  \end{split}
\end{equation}

Let $\Delta \ped_0^\tau = [\Delta x_0^\tau,\Delta y_0^\tau, \Delta \theta_z^\tau ] \in \mathbb R^3$ be the camera ego-motion from timestep $\tau-1$ to $\tau$ consisting of a 2D translation $[\Delta x_0, \Delta y_0]^\top$ and a change in viewing direction $\Delta \theta_z$ on the ground plane ($x$-$y$ plane). The optimal motion of the camera $\Delta \hat {\ped}^\tau_0$ and those of the pedestrians $\hat {\traj}^\tau_{1:K} = \{\ped_1^\tau, \ped_2^\tau, \dots , \ped^\tau_K\} \in \mathbb R^{2\times K}$ can be estimated as those that maximize the posterior distribution (\EQREF{eq:posterior}).
The motion of observed pedestrians $\traj_{1:K}^{\tau-1:\tau}$ are strictly constrained by the observing camera position $\ped_0^\tau$ and its viewing direction $\theta_z^\tau$.
With recovered pedestrian parameters $\hat {\traj}_{1:K}^\tau$, the optimal estimate of the camera ego-motion $\Delta \hat \ped_0^\tau$ becomes
\begin{equation}
  \begin{split}
    \Delta \hat \ped_0^\tau =& \argmax_{\Delta \ped_0^\tau \in \mathbb R^3} p(\ped_0^\tau | \traj_0^{\tau_0:\tau-1}) \\
    &\prod_k p(\ped_k^\tau|\hat{\traj}_k^{\tau_0:\tau-1}, \Delta \ped_0^\tau)p(\bm z_k^\tau | \bm x_k^\tau, \Delta \ped_0^{\tau})\,,
    \label{eq:optimal-x0}
  \end{split}
\end{equation}
where $p(\ped_0^\tau | \traj_0^{\tau_0:\tau-1})$ and $p(\ped_k^\tau | \traj_k^{\tau_0:\tau-1}, \Delta \ped_0^\tau)$ are motion priors of the camera and pedestrians conditioned on the camera motion, respectively. If the observer camera is mounted on a pedestrian following the crowd flow, $p(\ped_0^\tau|\traj_0^{\tau_0:\tau-1})$ obeys the same motion model as $p(\ped_k^\tau|\traj_k^{\tau_0:\tau-1})$.

As in previous work for pedestrian detection~\cite{luo2020whether}, we assume that the heights of pedestrians $h_k$ follow a Gaussian distribution. This lets us define the likelihood of observed pedestrian positions $\bm z_k^\tau$ relative to the camera $\ped_0^\tau$ as
\begin{equation}
    \bm z^\tau_k \sim p(\bm z^\tau_k |\ped_k^\tau;h_k)=\mathcal N(\mu_h,\sigma_h^2)\,,
    \label{eq:height-gaussian}
\end{equation}
where $\mathcal N(\mu_h,\sigma_h^2)$ is a Gaussian distribution with mean $\mu_h$ and variance $\sigma_h^2$.
Once the ego-motion of the observing camera is estimated as $\Delta \hat \ped_0^\tau$, the pedestrian positions $\hat \traj_{1:K}^\tau$ that maximize the posterior $p(\!\traj_{1:K}^\tau\!|\!\mathcal Z_{1:K}^\tau, \!\Delta \ped_0^\tau\!)$ can be obtained as
\begin{equation}
    \hat \traj_{1:K}^\tau = \argmax_{\ped_k^\tau \in \traj_{1:K}^\tau} \prod_k p(\ped_k^\tau|\traj_k^{\tau_0:\tau-1}, \Delta \hat \ped_0^\tau)p(\bm z_k^\tau | \bm x_k^\tau, \Delta \hat \ped_0^{\tau})\,.
    \label{eq:optimal-xk}
\end{equation}
That is, we can estimate the ego-motion of the observer constrained by the perceived pedestrian movements which conform to the crowd motion prior and the observation model.

When the camera observes a large number of pedestrians that conforms to a known crowd motion model, regarless of whether the camera motion is consistent with dominant crowd flow, the camera ego-motion estimates depend heavily on the observed crowd movements and are less sensitive to the assumed ego-motion model. In such cases, \EQREF{eq:optimal-x0} can be re-written as
\begin{equation}
  \begin{split}
    \Delta \hat \ped_0^\tau = \argmax_{\Delta \ped_0^\tau \in \mathbb R^3} \prod_{k=1}^{K} p(\ped_k^\tau|\hat{\traj}_k^{\tau_0:\tau-1}, \Delta \ped_0^\tau)p(\bm z_k^\tau | \bm x_k^\tau, \Delta \ped_0^{\tau})\,.
    \label{eq:optimal-x0-unknown-ego}
  \end{split}
\end{equation}
As long as the camera observes a sufficient number of pedestrians walking in diverse directions, our method can successfully birdify its view.

\subsection{Energy Minimization} \label{sec:energy-minimization}

Once the camera ego-motion is estimated, we can update the individual locations of pedestrians given the ego-motion in an iterative refinement process. View birdification can thus be solved with a cascaded optimization which first estimates the camera ego-motion and then recovers the relative locations between the camera and the pedestrians given the ego-motion estimate while taking into account the local interactions between pedestrians. Minimization of the negative log probabilities, \EQSREF{eq:optimal-x0} and \eqref{eq:optimal-xk}, can be expressed as
\begin{equation}
  \underset{\Delta \ped_0^\tau \in \mathbb R^3}{\text{minimize}}\,\,\,\ENERGY_c (
  \Delta \ped_0^\tau; {\hat \traj}^{\tau}_{1:K},{\mathcal Z}_{1:K}^\tau, \traj_{0:K}^{\tau_0:\tau-1})\,, \label{eq:energy-Ec}
\end{equation}
\begin{equation}
\begin{split}
  &\text{subject to} \\
  &\hat {\traj}_{1:K}^\tau = \underset{ \traj_{1:K}^\tau}{\text{argmin}}\, \ENERGY_p(\traj_{1:K}^\tau;\Delta \hat \ped_0^\tau, \mathcal Z_{1:K}^\tau, {\traj}_{0:K}^{\tau_0:\tau-1})\,, \label{eq:energy-Ep}
\end{split}
\end{equation}
where we define the energy functions for positions of camera $\ENERGY_c$ and pedestrians $\ENERGY_c$ as
\begin{equation}
\begin{split}
  \ENERGY_c (
  \Delta \ped_0^\tau;  {\hat \traj}^\tau_{1:K},{\mathcal Z}_{1:K}^\tau, \traj_{1:K}^{\tau_0:\tau-1})
  =-\mathrm{ln}\,p(\ped_0^\tau|\traj_0^{\tau_0:\tau-1}) + \ENERGY_p\,,\\
\end{split}
\end{equation}
\vspace{-12pt}
\begin{equation}
\begin{split}
  &\ENERGY_p(\traj_{1:K}^\tau;\Delta \hat \ped_0^\tau, \mathcal Z_{1:K}^\tau, {\traj}_{0:K}^{\tau_0:\tau-1})\\
  &= \sum_{k=1}^K \!\!\!-\mathrm{ln}\,p(\ped_k^\tau|\traj_k^{\tau_0:\tau-1},\Delta \ped_0^\tau) + \sum_{k=1}^K \!\!\!- \mathrm{ln}\,p(\bm z_k^\tau|\bm x_k^\tau,\Delta \ped_0^\tau)\,.
  \label{eq:energy-Ep-nlog}
\end{split}
\end{equation}

We minimize the energy in \EQREF{eq:energy-Ec} by first computing an optimal camera position $\hat \ped_0^\tau$ from \EQREF{eq:energy-Ec} with gradient descent and initial state $\ped_0^{\tau} = \ped_0^{\tau-1}$. Given the estimate of the observer location $\hat \ped_0^\tau$, we then estimate the pedestrian locations by solving the combinatorial optimization problem in \EQREF{eq:energy-Ep} for $\mathcal X_k^\tau$ while considering all possible combinations of $\{\ped_1^\tau, \dots, \ped_K^\tau \}$ under the projection constraint in \EQREF{eq:inv-projection} and the assumed pedestrian interaction model.

This can be interpreted as a fully connected graph consisting of $K$ pedestrian nodes with unary potential and interaction edges with pairwise potential.
Similar to prior works on low-level vision problems~\cite{badrinarayanan2014mixture,lezamacvpr2011}, \EQREF{eq:energy-Ep-nlog} can be optimized by iterative message passing~\cite{felzenszwalb2006efficient} on the graph.
The possible states $\ped_i$ are uniformly sampled on the projection line around $\mu_h$ with interval $[\mu_h-\delta S/2, \mu_h+\delta S/2]$, where $S$ is a number of samples and $\delta=0.01$.
Considering only pairwise interactions and Gaussian potential, the complexity of the optimization is $\mathcal O(KS^2T)$, where $T$ is the number of iterations required for convergence.
In this paper, we use two types of analytical interaction models, ConstVel~\cite{scholler2020constant} and Social Force~\cite{helbing1995social}. In what follows, we provide a detailed derivation of energy functions.

\subsection{Pedestrian Interaction Models}\label{sec:ped-interaction-model}
We formulated view birdification as an iterative energy minimization problem that consists of a pedestrian interaction model $p(\bm x^\tau_k|\traj^{\tau_0:\tau-1}_k)$
and a likelihood $p(\bm z_k^\tau|\ped_k^\tau,\Delta \ped_0^\tau)$ defined by the geometric observation model with ambiguities arising from human height estimates (\EQREF{eq:height-gaussian}).
Our framework is not limited to a specific pedestrian interaction model, and any type of model that explains pedestrian interactions in a crowd can be incorporated. In the following, we consider two example models with a temporal window of $\Delta \tau =2$.

\paragraph{Constant Velocity}
ConstVel~\cite{scholler2020constant} is a simple yet effective model of pedestrian interactions in a crowd which simply linearly extrapolates future trajectories from the last two frames
\begin{equation}
    p(\ped^\tau_k|\traj^{\tau-2:\tau-1}_k) \sim  \exp \left [ -\| \ped^\tau_k - 2\ped^{\tau-1}_k + \ped^{\tau-2}_k \|^2 \right ]\,.
\end{equation}
The model is independent of other pedestrians and the overall pedestrian interaction model can be factorized as $ p(\traj^\tau_{1:K} | \traj^{\tau-2:\tau-1}_{1:K})  = \prod_{k=1}^K p(\ped_k^\tau|\traj^{\tau-2:\tau-1}_k)$. The energy model $\ENERGY_p$ is rewritten as
\begin{equation}
    \ENERGY_p = \sum_{k=1}^K-\mathrm{ln}\,p(\ped_k^\tau|\traj_k^{\tau-2:\tau-1}) + \sum_{k=1}^K -\mathrm{ln}\,p(\bm z_k^\tau | \ped_k^\tau, \Delta \ped_0^\tau)\,.
\end{equation}

\paragraph{Social Force}
The Social Force Model~\cite{helbing1995social} is a well-known physics-based model that simulates multi-agent interactions with reciprocal forces, which is widely used in crowd analysis and prediction studies~\cite{mehran2009abnormal,van2011reciprocal}. Each pedestrian $k$ with a mass $m_k$ follows the velocity $d\ped / dt^2$
\begin{equation}
    m_k \frac{d^2 \ped_k}{dt^2} = \bm F_{k} = \Fp(\ped_k) + \Fi(\mathcal X_{\mathcal C})\,,
\end{equation}
where $\bm F_k$ is the force on $\ped_k$ consisting of the personal desired force $\Fp$ and the reciprocal force $\Fi$. The personal desired force is proportional to the discrepancy between the current velocity and that desired
\begin{equation}
    \Fp(\ped_k) = \frac{1}{\eta}\left (\vpref_k-\frac{d\ped_k}{dt}\right ),
\end{equation}
where $\vpref_k$ denotes the desired velocity which can be empirically approximated as the average velocity of neighboring pedestrians $i \in \mathcal N(\ped_k)$~\cite{mehran2009abnormal}.

The form of reciprocal force $\Fi$ can be determined by the set of interactions between pedestrian nodes $\bm x_i \in \mathcal X_{\mathcal C}$. To reduce the complexity of optimization, we approximate multi-human interaction $\Fi(\mathcal X_{\mathcal C})$ with a collection of pairwise interactions $\Fi(\ped_i, \ped_k)$. We assume a standard Gaussian potential to simulate the reciprocal force between two pedestrians
\begin{equation}
    \Fi(\ped_i,\ped_k)\! = -\nabla \left( \frac{1}{\sqrt{2\pi\sigma^2}}\exp\left[-\frac{\|\ped_i - \ped_k \|^2}{2\sigma^2}\right]\!\right)\,.
    \label{eq:pairwise-potential}
\end{equation}

Without loss of generality, we omit $m_k$ as $m_k=1$, assuming that the mass of pedestrians in a crowd is almost consistent.
Taking the last two frames as inputs, the complete pedestrian interaction model becomes
\begin{equation}
\begin{split}
    & p(\traj^\tau_{1:K} | \traj^{\tau-2:\tau-1}_{1:K})\\
    &\!\!\sim \prod_k \exp \left[\!- \left \| \Fp(\ped_k^\tau) - \frac{d^2\ped_k^\tau}{dt^2} \right \|\right]\!\!\prod_{(i,k)\in \mathcal X_{\mathcal C}}\!\!\!\!\!\exp\left [- \left \| \Fi(\ped_i^\tau,\ped_k^\tau)\right \|\right ]\,.
\end{split}  \label{eq:sf-interaction-model}
\end{equation}

Taking negative log probabilities, the overall energy model in \EQREF{eq:sf-interaction-model} becomes
\begin{equation}
\ENERGY_{p} = \sum_k D_k(\ped_k^\tau;\mathcal X^{\tau-2:\tau-1}_k) + \!\!\!\sum_{(i,k)\in \mathcal X_{\mathcal C}}\!\!\!V_{ik}(\ped_i^\tau, \ped_k^\tau)\,,
\end{equation}
where the unary term and pairwise terms are
\begin{align}
    &D_k(\ped_k^\tau) = \left \|\Fp(\ped_k^\tau) - \frac{d^2\ped_k^\tau}{dt}\right \| - \mathrm{ln}\,p(\bm z_k^\tau | \ped_k^\tau, \Delta \ped_0^\tau)\,, \\
    &V_{ik}(\ped_i^\tau,\ped_k^\tau) = \Fi(\ped_i^\tau, \ped_k^\tau)\,,
\end{align}
respectively.

\subsection{Optimization over a Large Number of Pedestrians} \label{sec:opt-selective-pedestrians}
In highly congested environments (\eg, $K>100$), the computational cost for optimizing \EQREF{eq:energy-Ep} increases linearly in the number of pedestrians $K$.
To handle realistic scenarios in which most of the pedestrians in the crowd are occluded by others, we use $\tilde K$ selected pedestrians whose size is above a predetermined threshold $\epsilon$. We define a set of neighboring pedestrians at time $\tau$ 
$\mathcal N(\ped_0^\tau) = \{\ped_k^\tau : f_n(u_k,v_k) \geq \epsilon \}$, where $f_n(u_k,v_k)$ returns the size of the bounding boxes observed in the image.
The energy minimization for the neighboring pedestrians becomes
\begin{equation}
\hat{\mathcal N(\ped_0^\tau)} = \underset{k \in \{k: f_n(u_k^\tau, v_k^\tau) \geq \epsilon \}}{\text{argmin}}\, \ENERGY_p(\ped_k^\tau;\Delta \hat \ped_0^{\tau_0:\tau}, \bm z_k^\tau, {\ped}_k^{\tau_0:\tau-1})\,. \label{eq:energy-Ep2}
\end{equation}
Note that optimizing positions of only foreground pedestrians may result in inaccurate localization due to the incomplete interaction model that considers only a small part of the whole crowd. Nevertheless, in Sec. \ref{sec:vb-result}, we show that our proposed framework achieves sufficient localization accuracy even with the a small number of selected pedestrians in a super-dense crowd.

\subsection{Implementation Details}
We use the validation split of each crowd dataset~\cite{ivanovic2019trajectron} to find the optimal hyperparameters of the pedestrian interaction models.
We set the weight parameter of the desired force $\Fp$ to $\eta=0.5$, and the variance of the Gaussian potential to $\sigma^2=1.0$ for the social force model. For each dataset of simulated and real trajectories, the size of the ground field, where pedestrians are walking from starting points to their destinations, is scaled to [$-8.0,8.0$] m. We also assume that the initial positions of pedestrians $\ped^{\tau_1}_k$ and $\ped^{\tau_1+1}_k$ for time $\tau_1,\tau_1+1$ are given a priori, and the positions at the next timesteps $\traj_k^{\tau_1+2:\tau_2} = \{\ped^{\tau_1+2}_k, \dots , \ped^{\tau_2}_k\}$ are sequentially estimated based on our approach.

\section{Experiments}
\def\CB{{\small\CheckmarkBold}}
\begin{table*}[t]
  \center
  \caption{\textbf{Overview of birdification dataset.} For real trajectories, we selected scenes of Hotel, ETH, and Students by taking into account the number of people in the crowd. ``Seq." corresponds to all the frames captured by a moving observer. ``Len." denotes the number of frames included in one sequence.}
  \label{table:dataset}
  \vspace{0.5\baselineskip}
  \small
  \begin{tabularx}{1.0\linewidth}{ccccccccccc}
    \toprule[1pt]
    \multirow{2}{*}{\textbf{Dataset}} & \!\!\!\textbf{Seq.} & \textbf{Len.} & \multicolumn{3}{c}{\textbf{People in Crowd}} & \textbf{Int.} & \textbf{observer} & \textbf{input} & \textbf{height} & \textbf{occluded} \\
    &\!\!\!Total&Avg&Min&Avg&Max & \textbf{model} & \textbf{view} & \textbf{bboxes} & \textbf{variances} & \textbf{pedestrians} \\
    \toprule[1pt]
        \textbf{Sim} & 500  & 20.0 & 10 & --- & 50 & synthetic & synthetic& given &\CB & \\
        \midrule
\textbf{Hotel} & 340 & 15.0&3 & 6.31 & 15 & real & synthetic & given & \CB \\
\textbf{ETH} & 346 &14.4& 3 & 9.29& 26 & real & synthetic & given & \CB\\
\textbf{Students} & 849 &45.8& 13 & 44.2& 75 & real & synthetic & given &\CB\\
\textbf{Shibuya01} & 806 & 317 & 1 & 523 & 770 & real & synthetic & given & \CB \\ 
\textbf{Shibuya02}  & 568 & 299 & 25 &281 & 492 & real & synthetic & given & \CB  

\\
    \midrule
    \textbf{GTAV} &--- & 400 & 3 & 6 & 12 & synthetic & photorealistic & MOT~\cite{wang2019towards} &  & \CB\\
    \bottomrule
  \end{tabularx}
\end{table*}

\begin{figure*}[t]
    \centering
    \includegraphics[width=1.0\linewidth]{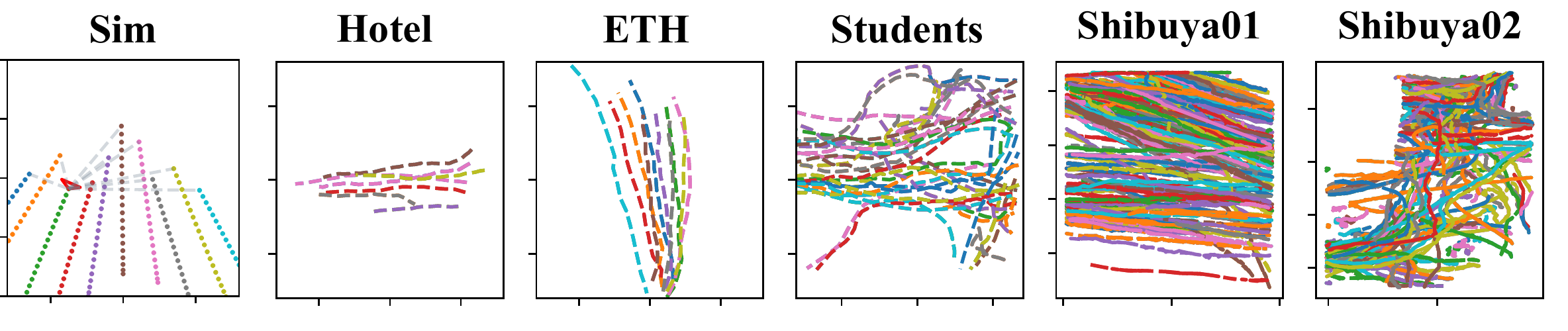}
    \caption{\textbf{Typical example trajectories.} Typical example trajectories from the datasets Sim, Hotel, ETH, Students, and Shibuya.
    In the Sim Example, the red triangle is the virtual camera that observes projected pedestrians on the image plane, where dashed gray lines denote the projection.}
    \label{fig:crowd_example}
\end{figure*}
\begin{figure}[t]
    \centering
    \includegraphics[width=\linewidth]{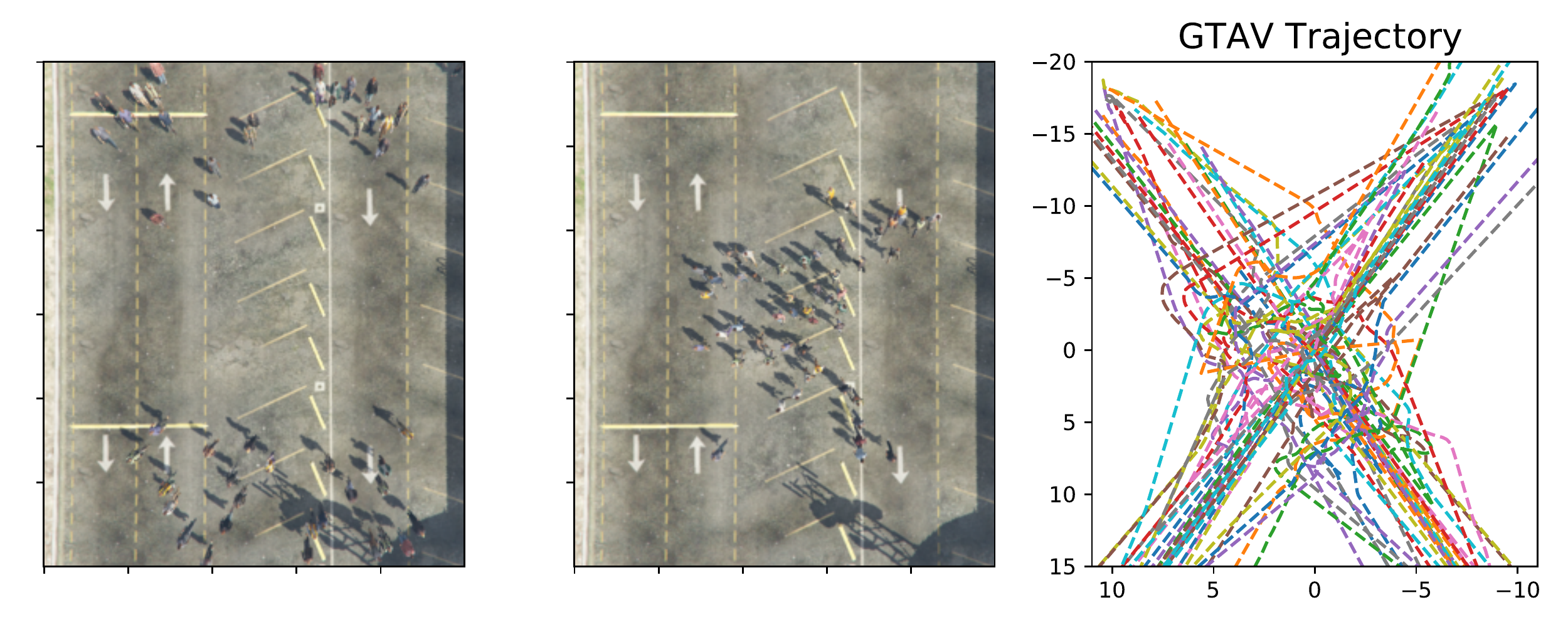}
    \caption{\textbf{Example Trajectories from the GTAV dataset.} (Left) Pedestrians are spawned at one of the four corners of the field. (Center) Pedestrians walking towards their destinations while avoiding collisions. (Right) Trajectories of each pedestrian in one sequence.}
    \label{fig:crowd_example_gta}
\end{figure}

We validate the effectiveness of the proposed geometric view birdification method through an extensive set of experiments. We constructed several datasets consisting of synthetic pedestrian trajectories (\textbf{Sim}), real pedestrian trajectories (\textbf{Hotel}, \textbf{ETH}, \textbf{Students}, and \textbf{Shibuya}), and photorealistic crowd simulation (\textbf{GTAV}).
These datasets differ in several aspects (\ie, density of crowd, synthetic view or not, synthetic or real interaction models).
Table \ref{table:dataset} summarizes the statistics and taxonomy of these datasets.
We also validate our method on a real mobile robot-view dataset~\cite{martin2021jrdb} consisting of a pair of real $360^\circ$ cylindrical images and 2D-3D bounding box annotations of surrounding pedestrians.

\subsection{View Birdification Datasets}
To the best of our knowledge, no public dataset is available for evaluating view birdification (\ie, ego-video in crowds). We construct the following three datasets, which we will publicly disseminate, for evaluating our method and also to serve as a platform for further studies on view birdification.

\paragraph{Synthetic Pedestrian Trajectories}
The first dataset consists of synthetic trajectories paired with their synthetic projections to an observation camera. This data allows us to evaluate the effectiveness of view birdification when the crowd interaction model is known. The trajectories are generated by the social force model~\cite{helbing1995social} with a varying number of pedestrians $K \in \{10,20,30,40,50\}$, and a perspective observation camera mounted on one of them. To evaluate the validity of our geometric formulation and optimization solution with this dataset, we assume ideal observation of pedestrians, \ie, pedestrians do not occlude each other and their projected heights can be accurately deduced from the observed images. We also assume that the pedestrians are extracted from the ego-centric video perfectly but their heights $h_k$ are sampled from a Gaussian distribution $h_k \sim \mathcal N(\mu_h,\sigma_h^2)$ with mean $\mu_h = 1.70$ [m] and a standard deviation $\sigma_h \in [0.00,0.07]$ [m] based on the statistics of European adults~\cite{visscher2008}.

\paragraph{Real Pedestrian Trajectories}
The second dataset consists of real pedestrian trajectories paired with their synthetic projections on an observation camera's image-plane. The trajectories are extracted from publicly available crowd datasets: three sets of sequences referred to as \textbf{Hotel}, \textbf{ETH}, and \textbf{Students} are from ETH~\cite{pellegrini2009you} and UCY~\cite{lerner2007crowds}. The two referred to as \textbf{Shibuya01} and \textbf{Shibuya02} are from CroHD dataset~\cite{sundararaman2021tracking}.
As in the synthetic pedestrian trajectories dataset, we render corresponding ego-centric videos from a randomly selected pedestrian's vantage point.
Hotel, ETH, Students, and Shibuya datasets correspond to sparsely, moderately, densely, and super-densely crowded scenarios, respectively. This dataset allows us to evaluate the effectiveness of our method on real data movements.

\paragraph{Photorealistic Crowd Simulation} The third dataset consists of synthetic trajectories paired with their photo-realistic projection captured with the limited field of views and frequent occlusions between pedestrians. Evaluation on this dataset lets us examine the end-to-end effectiveness of our method including robustness to tracking errors. Inspired by previous works on crowd analysis and trajectory prediction~\cite{wang2019learning,caoHMP2020}, we use the video game engine of \emph{Grand Theft Auto V} (GTAV) developed by \emph{Rockstar North}~\cite{rockstar} with crowd flows automatically generated from programmed destinations with collision avoidance. We collected pairs of ego-centric videos with $90^\circ$ field-of-view and corresponding ground truth trajectories on the ground plane using Script Hook V API~\cite{scripthookv}. We randomly picked $50$ different person models with different skin colors, body shapes, and clothes. We prepare two versions of this data, one with manually annotated centerline and heights of the pedestrians in the observed video frames and the other with those automatically extracted with a pedestrian detector~\cite{wang2019towards} pretrained on MOT-16~\cite{milan2016mot16} which includes data captured from a moving platform.

\subsection{Example Trajectories}
\FIGREF{fig:crowd_example} visualizes typical example sequences from the synthetic dataset referred to as \textbf{Sim} and from the real trajectory dataset referred to as \textbf{Hotel}, \textbf{ETH}, \textbf{Students}, and \textbf{Shibuya}.
In all of these datasets, a virtual observation camera is assigned to one of the trajectories and the observer captures the rest of the pedestrians in the sequence.
\FIGREF{fig:crowd_example_gta} shows example trajectories of the GTAV dataset. The size of the ground field, where pedestrians are walking from starting points to their destinations, is configured to be $20{\text{m}}\times 40{\text{m}}$.
We spawned $50$ pedestrians starting from one of the four corners of the field, $[-10,-10], [10,10], [10,-20], [10,20]$, and set the opposite side of the field as their destinations. Both the starting points and destinations were randomized with a uniform distribution.
In the \textbf{GTAV} dataset, an observation camera is mounted on one of the pedestrians walking in the crowd flow and we can obtain pairs of ground-truth trajectories and ego-centric videos with $90^\circ$ field -of-view via Script Hook V APIs~\cite{scripthookv}.

\subsection{View Birdification Results}\label{sec:vb-result}

\paragraph{Evaluation Metric.}
We quantify the accuracy of our method by measuring the differences between the estimated positions of the pedestrians $\ped_k^\tau$ and the observer $R^\tau, \ped_0^\tau$ on the ground plane from their ground truth values $\Xgt_k^\tau, \Rgt^\tau$, and $\Xgt_0^\tau$, respectively. The translation error for the observer is $\Terr=\frac{1}{T}\sum^\tau\|{\ped_0}^\tau - \Xgt_0^\tau\|$, where $\tau$ is a timestep duration of the sequence. The rotation error of the observer is $\Rerr =\frac{1}{T}\sum_t \mathrm{arccos}(\frac{1}{2}~\mathrm{trace}( R^\tau(\Rgt^\tau)^\top-1)$. We also evaluate the absolute and relative reconstruction errors of surrounding pedestrians which are defined by $\Xerr= \frac{1}{K}\frac{1}{T}\sum_k \sum_t \| \bm { x}_k^\tau - \Xgt^\tau_k\|\,$ and $\XerrRel = \frac{1}{K}\frac{1}{T}\sum_k\sum_t \|( {\bm x}_k^\tau -{\bm x}_0^\tau) - ( \Xgt_k^\tau - \Xgt_0^\tau) \|\,$, respectively.
\begin{table*}[t]
  \centering
  \caption{\textbf{Birdification results on real trajectories.} Relative and absolute localization errors of pedestrians, $\Delta \tilde \ped, \Delta \ped$ (top), and camera ego-motion errors, $\Delta r$ and $\Delta \bm t$ (bottom), were computed for each frame for three different video sequences. Baseline methods only extrapolate movements on the ground plane resulting in missing entries (--). The results demonstrate the effectiveness of our view birdification.}
  \label{table:quant-ped-localization}
  \vspace{0.5\baselineskip}
  \small
  \begin{tabularx}{1.0\linewidth}{cccccccc}
    \toprule[1pt]
     \multirow{2}{*}{Dataset} & &\multicolumn{2}{c}{\textbf{Hotel / sparse}} & \multicolumn{2}{c}{\textbf{ETH / mid}} & \multicolumn{2}{c}{\textbf{Students / dense}} \\
     &$\sigma_h$ & $\XerrRel$ [m]&  $\Xerr$ [m]&$\XerrRel$ [m]&  $\Xerr$ [m]& $\XerrRel$ [m]&  $\Xerr$ [m]\\
    \toprule[1pt]
        CV~\cite{scholler2020constant} &-- & -- & 0.294 $\pm$ 0.186 & -- & 0.275 $\pm$ 0.195 & -- & 0.223 $\pm$ 0.169\\
        SF~\cite{helbing1995social} &-- & -- & 0.289 $\pm$ 0.207 & -- & 0.261 $\pm$ 0.174 & -- & 0.222 $\pm$ 0.163\\
    \midrule
        \textbf{\proposed-CV} & 0.00 & 0.051 $\pm$ 0.029 & 0.070 $\pm$ 0.030 & 0.089 $\pm$ 0.045 & 0.115 $\pm$ 0.049 & 0.022 $\pm$ 0.008 & 0.023 $\pm$ 0.008\\
        & 0.07 &0.051 $\pm$ 0.029 & 0.070 $\pm$ 0.030 & 0.090 $\pm$ 0.045 & 0.116 $\pm$ 0.050 & 0.021 $\pm$ 0.007 & 0.022 $\pm$ 0.008\\
        \textbf{\proposed-SF} & 0.00 & \textbf{0.048 $\pm$ 0.027} & \textbf{0.052 $\pm$ 0.033} & \textbf{0.070 $\pm$ 0.040} & \textbf{0.079 $\pm$ 0.047} & \textbf{0.009 $\pm$ 0.003} & \textbf{0.010 $\pm$ 0.006}\\
        & 0.07 & \textbf{0.049 $\pm$ 0.027} & \textbf{0.052 $\pm$ 0.032} & \textbf{0.071 $\pm$ 0.040} & \textbf{0.080 $\pm$ 0.047} & \textbf{0.009 $\pm$ 0.004} & \textbf{0.010 $\pm$ 0.006} \\
    \midrule[0.3pt]\midrule[1pt]
         &$\sigma_h$ & $\Rerr$ [rad]&  $\Terr$ [m]&$\Rerr$ [rad]& $\Terr$ [m] &$\Rerr$ [rad]& $\Terr$[m]\\
    \toprule[1pt]
        \textbf{\proposed-CV}
        & 0.00 & \textbf{0.015 $\pm$ 0.030} & 0.066 $\pm$ 0.089 & 0.016 $\pm$ 0.027 & 0.095 $\pm$ 0.125 & \textbf{0.001 $\pm$ 0.001} & 0.010 $\pm$ 0.007\\
        & 0.07 & 0.017 $\pm$ 0.039 & 0.069 $\pm$ 0.100 & 0.019 $\pm$ 0.034 & 0.110 $\pm$ 0.148 & \textbf{0.001 $\pm$ 0.001} & 0.010 $\pm$ 0.007\\
        \textbf{\proposed-SF} & 0.00 & 0.015 $\pm$ 0.036 & \textbf{0.062 $\pm$ 0.104} & \textbf{0.015 $\pm$ 0.031} & \textbf{0.089 $\pm$ 0.135} & \textbf{0.001 $\pm$ 0.001} & \textbf{0.009 $\pm$ 0.006}\\
        & 0.07 & \textbf{0.016 $\pm$ 0.042} & \textbf{0.062 $\pm$ 0.103} & \textbf{0.016 $\pm$ 0.035} & \textbf{0.091 $\pm$ 0.153} & \textbf{0.001 $\pm$ 0.001} & \textbf{0.009 $\pm$ 0.006}\\
     \bottomrule
  \end{tabularx}
\end{table*}

\begin{figure}[t]
    \centering
    \includegraphics[width=1.0\linewidth]{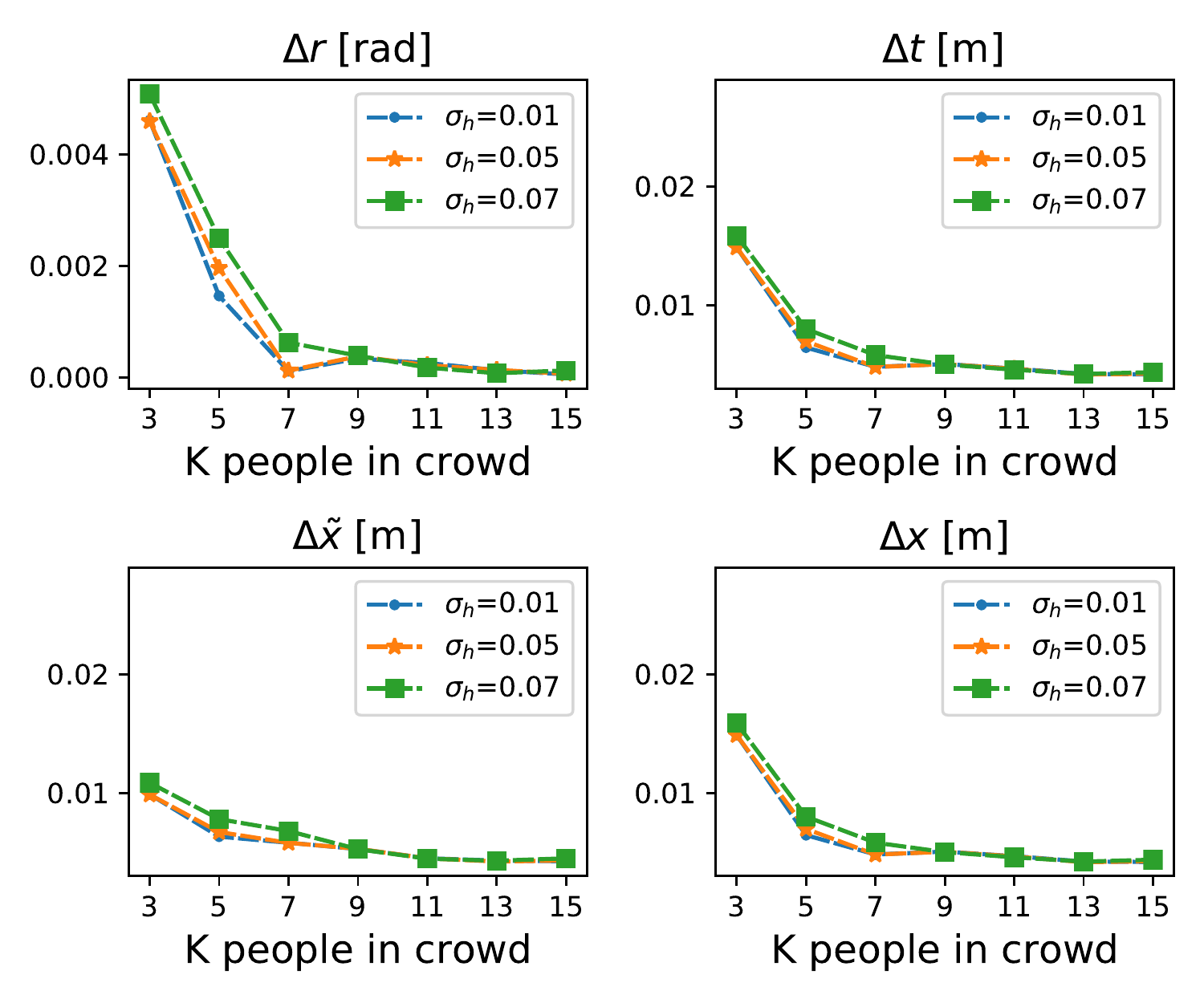}
    \caption{\textbf{Results on synthetic pedestrian trajectories.} Circle, star, and squared markers denote errors of estimated camera rotations $\Delta r$, translations $\Delta \bm t$, relative $\Delta {\tilde{\bm x}}$ and absolute localization errors $\Delta {\bm x}$,  respectively, with standard deviations of pedestrian heights, $\sigma=0.01,0.05,0.07$ [m], respectively.}
    \label{fig:sim_sigma_result}
\end{figure}

\begin{table}[t]
  \centering
  \caption{\textbf{Birdification results in the super-dense crowd.} Relative and absolute localization errors of pedestrians, $\Delta \tilde \ped, \Delta \ped$ (top), and camera ego-motion errors, $\Delta r$ and $\Delta \bm t$ (bottom), were computed for each frame for three different video sequences. Baseline methods only extrapolate movements on the ground plane resulting in missing entries (--). The results demonstrate the effectiveness of our view birdification even in super-dense crowds.}
  \label{table:quant-super-dense}
  \vspace{0.5\baselineskip}
  \small
  \begin{tabularx}{1.0\linewidth}{lccccc}
    \toprule[1pt]
     \multirow{2}{*}{Dataset} & &\multicolumn{2}{c}{\textbf{Shibuya01}} & \multicolumn{2}{c}{\textbf{Shibura02}} \\
     &$\sigma_h$ & $\XerrRel$ [m]&  $\Xerr$ [m]&$\XerrRel$ [m]&  $\Xerr$ [m] \\
    \toprule[1pt]
        CV~\cite{scholler2020constant} &-- & -- & 0.221 & -- & 0.245 \\
        SF~\cite{helbing1995social} &-- & -- & 0.220 & -- & 0.249 \\
    \midrule
        \textbf{\proposed-CV} & 0.07 & 0.023 & 0.024 & 0.025 & 0.026 \\
        \textbf{\proposed-SF} & 0.07 & 0.022 & 0.023 & 0.024 & 0.025 \\
    \midrule[0.3pt]\midrule[1pt]
     &$\sigma_h$ & $\Rerr$ [rad]&  $\Terr$ [m]&$\Rerr$ [rad]& $\Terr$ [m] \\
    \toprule[1pt]
        \textbf{\proposed-CV} & 0.07 & 0.001 & 0.011 & 0.001 & 0.012 \\
        \textbf{\proposed-SF} & 0.07 & 0.001 & 0.011 & 0.001 & 0.011 \\
     \bottomrule
  \end{tabularx}
\end{table}

\paragraph{Results on Known Interaction Model.}
\FIGREF{fig:sim_sigma_result} shows the view birdification results on the synthetic trajectories dataset.
Although both rotation and translation errors slightly increase as the height standard deviation $\sigma_h$ becomes larger, the error rate becomes lower as the number of people $K$ increases. This suggests that the more crowded, the more certain the camera position and thus the more accurate the birdification of surrounding pedestrians.

\begin{figure*}[t]
\centering
\includegraphics[width=0.9\linewidth]{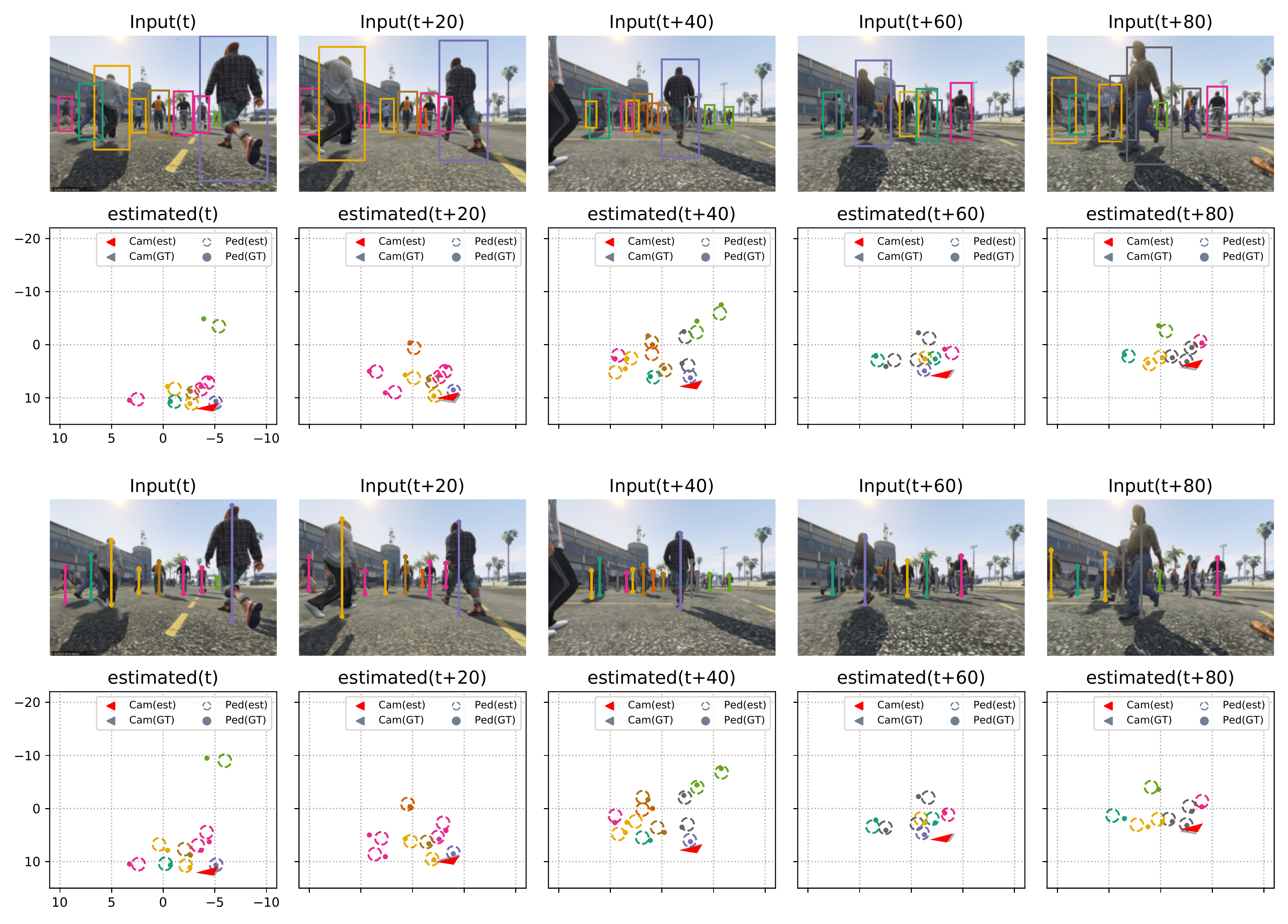}
    \caption{\textbf{Results on photorealistic crowd dataset.}
The top row shows detected pedestrians with a multi-object tracker in bounding boxes and the third row shows manually annotated human heights (center lines). The figures in the second and fourth rows depict view birdification results for them. Colors correspond to Pedestrian IDs. Red triangles denote camera position estimates $\ped_0^\tau$ and dashed circles denote estimated pedestrian positions $\ped_k^\tau$ at time $\tau$. Grey triangles and circles denote ground-truth camera and pedestrian positions, respectively. View birdificaiton results for both automatic and manually detected people show consistently high accuracy. These results demonstrate the end-to-end accuracy of view birdification.}
    \label{fig:gta5_concat}
\end{figure*}

\paragraph{Results on Unknown Real Interaction Models.}
The real trajectories data allow us to evaluate the accuracy of our method when the interactions between pedestrians are not known. We employ two pedestrian interaction models, Social Force~(SF)~\cite{helbing1995social} and ConstVel~(CV)~\cite{scholler2020constant}. We first evaluate the accuracy of our view birdification (VB) using these models, referred to as \emph{\proposed-SF} and \emph{\proposed-CV}, and compare them with baseline prediction models. In these baseline models, referred to as \emph{ConstVel~(CV)} and \emph{Social Force~(SF)}, we extrapolate a pedestrian position $\traj^\tau_{k}$ from its past locations $\mathcal X^{\tau-2:\tau-1}_{k}$ based on the corresponding interaction model without using the observer's ego-centric view. That is, the baseline model is not view birdification but extrapolation according to pre-defined motion models on the ground plane.

Table \ref{table:quant-ped-localization} shows the errors of our method and baseline models. These results clearly show that our method, both \emph{\proposed-CV} and \emph{\proposed-SF}, can estimate the camera ego-motion and localize surrounding people more accurately, which demonstrates the effectiveness of birdifying the view and exploiting the geometric constraints on the pedestrians through it. \emph{\proposed-SF} performs better than \emph{\proposed-CV} especially in scenes with rich interactions such as ETH and Students, while they show similar performance on the Hotel dataset that includes less interactions. Both \emph{\proposed-SF} and \emph{\proposed-CV} show accurate camera ego-motion results in the Students dataset, which demonstrates the robustness of ego-centric view localization regardless of the assumed pedestrian interaction models. Our method achieves high accuracy on all three datasets across different standard deviations of heights $\sigma_h \in [0.00, 0.07]$. This also shows that the method is robust to variation in human heights.

\paragraph{Selecting pedestrians in Super Dense Crowds.}
Table \ref{table:quant-super-dense} shows the localization errors of the view birdification and the baseline models on the super-dense crowd datasets (Shibuya01, Shibuya02).
In a highly congested scenario $K>100$, we can no longer consider all the pedestrians due to computational cost.
Following Sec.\ref{sec:opt-selective-pedestrians}, for each frame we select a set of  pedestrians in the neighborhood of the camera $\mathcal N(\ped_0)$ with a threshold function $f_n(u_k,v_k) = \|u_k\cdot v_k\| \geq \epsilon$ and $\epsilon =10.0$.
Even if we do not consider all the pedestrians in the crowd for optimizing with the assumed interaction model, our model \emph{VB-CV} and \emph{VB-SF} still demonstrate comparable localization accuracy with those for the Students dataset.
This is because our assumed interaction model considers up to first-order interactions with other pedestrians in close proximity, and the interaction models with selected neighboring pedestrians can approximate the whole interaction models in super dense crowd with sufficient accuracy.
These results clearly demonstrate our model can be applied to highly dynamic crowded environments, where static keypoints-based SLAM fails.

\begin{table}[t]
  \centering
  \caption{\textbf{Quantitative Results on GTAV dataset for different inputs.} The relative and absolute localization errors of pedestrians, $\Delta \tilde \ped$ and $\Delta \ped$, respectively, and the errors of camera ego-motion estimation, $\Delta r$, and $\Delta \bm t$, computed for each frame whose mean values are shown.
  \textbf{cline} denotes ideal, manually annotated inputs and \textbf{MOT} denotes inputs with detection noise by multi-object tracker~\cite{wang2019towards}.}
  \label{tab:quant-gtav}
  \vspace{0.5\baselineskip}
  \small
  \begin{tabularx}{\linewidth}{ccccc}
    \toprule[1pt]
     Input & $\Rerr$ [rad]&  $\Terr$ [m]&$\XerrRel$ [m]&  $\Xerr$ [m] \\
    \midrule
        \textbf{cline(manual)} & 0.015 & 0.097 & 0.441 & 0.491\\
        \textbf{MOT}~\cite{wang2019towards} & 0.016 & 0.101 & 0.491 & 0.530\\
    \toprule[1pt]
  \end{tabularx}
\end{table}

\begin{figure*}[t]
    \centering
        \includegraphics[width=0.95\linewidth]{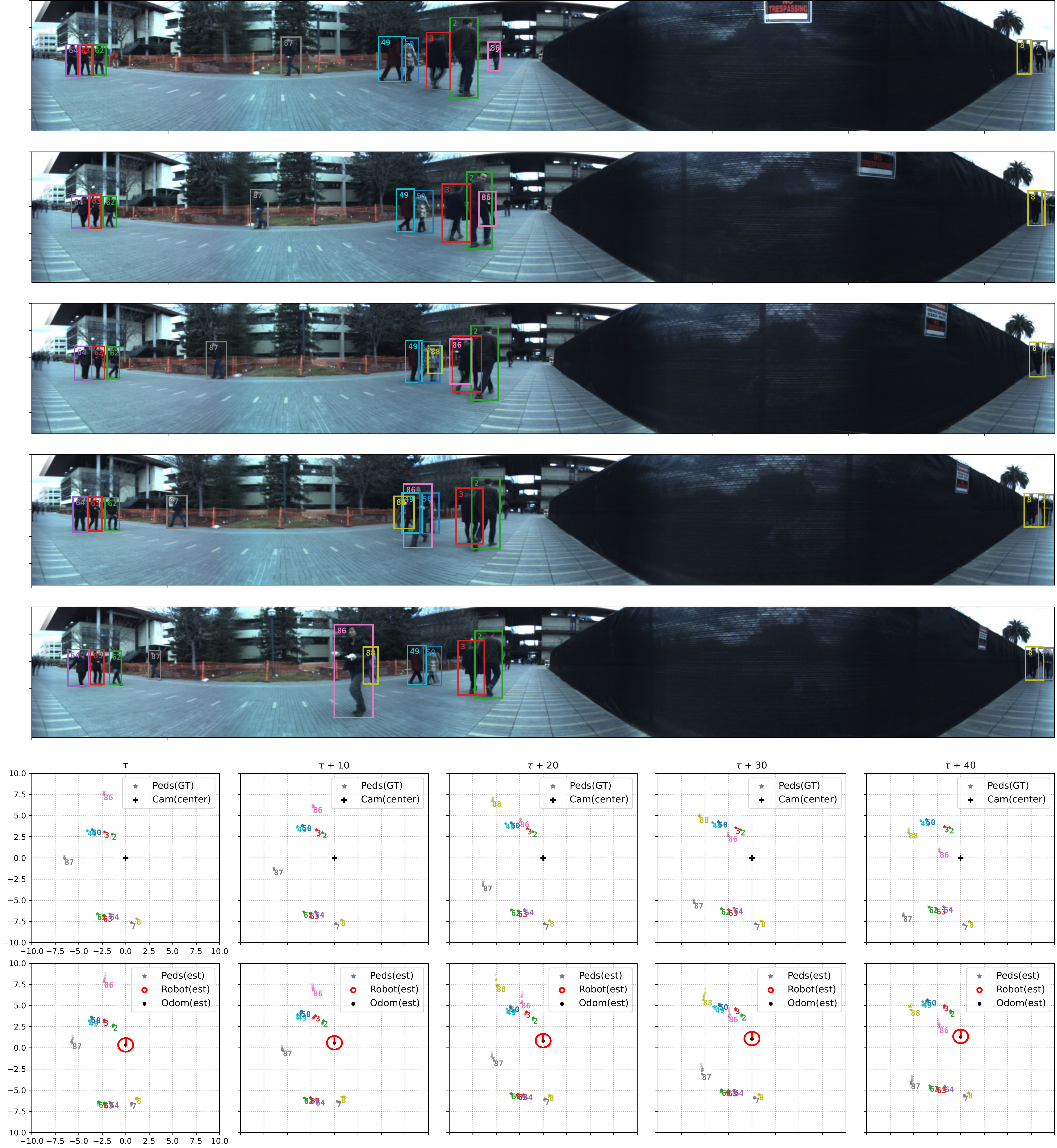}
    \caption{\textbf{Results on real robot dataset.} The top four rows show 2D bounding box annotations for pedestrians in the cylindrical RGB image at $\tau \in (\tau, \tau+10,\tau+20,\tau+30,\tau+40)$. The fourth row depicts ground-truth global layout of pedestrians relative to the camera $\ped_0 = [0,0]^\top$ at every timestep. The fifth row shows the view birdification results given the sequence of 2D bounding box movements in the cylindrical RGB images. Colors correspond to pedestrian IDs. Red triangles denote camera position estimates. }
    \label{fig:jrdb}
\end{figure*}

\paragraph{Photorealistic Crowds.}
\FIGREF{fig:gta5_concat} shows qualitative results on the photorealistic crowd dataset.
We prepared two versions of inputs, one manually annotated with centerlines of the people and their heights and the other with those automatically extracted from a multi-object tracker (MOT).
We compared view birdification results using these two different inputs, which are referred to as \emph{VB-cLine} and \emph{VB-MOT}.
 As shown in the top two rows, \emph{VB-MOT} accurately estimates camera ego-motion and on-ground positions of automatically detected pedestrians with an off-the-shelf tracker~\cite{wang2019towards}. People tracked in more than three frames are birdified. Even with occlusions in the image and noisy height estimates computed from detected bounding boxes, our approach robustly estimates the camera ego-motion and surrounding pedestrian positions. Due to perspective projection, localization error caused by erroneous detection in the image plane is proportional to the ground-plane distance between the camera and the detected pedestrian. We further compared these results with \emph{VB-cline} as shown in the bottom two rows \FIGREF{fig:gta5_concat} to highlight the effect of automatically detecting the pedestrians for view birdification (\ie, to see how the results change if the pedestrian heights were accurate). The resulting accuracies are comparable, which demonstrates the end-to-end effectiveness.
To further ameliorate the errors caused by detection noises, our method can also be extended, for instance, by replacing the noise model in \EQREF{eq:height-gaussian} with a 2D Gaussian distribution.

\subsection{Unknown Ego-Motion Recovery with the Real Mobile Platform Dataset}
\paragraph{JackRabbot Dataset.} 
We also test on the JackRabbot Dataset and Benchmarks (JRDB)~\cite{martin2021jrdb}, which includes panorama ($360^\circ$) RGB images with 2D-3D bounding box annotations of pedestrians captured by a mobile robot platform of human-compatible size.
The robot captures the social interactions of a crowd in outdoor/indoor environments, where all the pedestrian IDs are assigned and their 3D locations are annotated in the relative coordinate system of the mounted camera.
The camera ego-motion is constrained to the 2D motion on the ground, \ie, $R\in SO(2)$, $\bm t \in \mathbb R^2$. 
The notable difference from our view birdification datasets is, the motion model of the ego-motion does not conform with the crowd motion model of surrounding pedestrians. 
This dataset allows us to evaluate the applicability of our method on mobile robot platforms with unknown motion model.
In this dataset, we use the cylindrical projection model described in \EQREF{eq:cylindrical-projection} for $360^\circ$ cylindrical RGB image inputs, and reconstruct both the ego-motion and pedestrian trajectories in absolute coordinates only from observed 2D movements in the image.

\paragraph{Comparison with the robot localization results from sensor values.}
We compare the localization results with that estimated from IMU sensor values and the wheel odometry recorded in the rosbag of the dataset.
As no ground-truth ego-motion is available for this dataset, we create pseudo localization results by fusing these sensor values with an extended Kalman Filter~\cite{MooreStouchKeneralizedEkf2014}.
\FIGREF{fig:jrdb} demonstrates our view birdification results with an unknown ego-motion model.
Our method can successfully recover the on-ground absolute trajectories of both the camera and its surrounding pedestrians. Even in these scenarios in which the camera ego-motion model is not consistent with the assumed crowd motion model (\eg, a mobile robot platform), our method can recover the camera ego-motion as long as the camera observes the pedestrians with an assumed motion model.
Both sensor-based and our vision-based localization have uncertainties arising from the observation errors, which often results in a significant ego-motion drift in long-term navigation. Even if the mobile platform is equipped with an IMU and other odometry sensors, the birdification results are still essential for obtaining reliable ego-motion estimates and can provide a reliable source for sensor fusion.

\begin{figure*}[t]
    \centering
    \includegraphics[width=\linewidth]{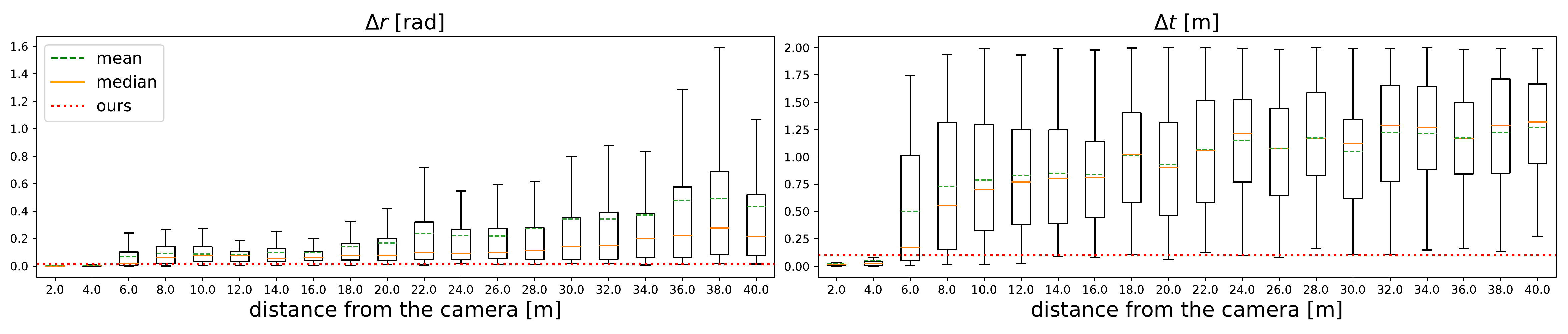}
    \caption{\textbf{Robustness of relative pose solver~\cite{nister2004efficient} against noise according to the distances from keypoints.}
These two figures show errors in pose estimation consisting of rotation $\Delta r$ [rad] and translation $\Delta \bm t$ [m].
The boxes indicate the range between 25th and 75th percentiles from the lowest values, where the orange and green dashed lines medians and means of the errors, respectively.
The black whiskers extends from the lowest to highest values. The red dotted line indicates the error of our method with MOT input noise (Table \ref{tab:quant-gtav}).
While the geometric solver works well with keypoints captured at $\leq 2$ [m], the accuracy of relative pose estimation significantly drops when captured at $\geq 6$ [m].}
    \label{fig:static-vs-dynamic}
\end{figure*}

\begin{figure}[t]
    \centering
    \includegraphics[width=\linewidth]{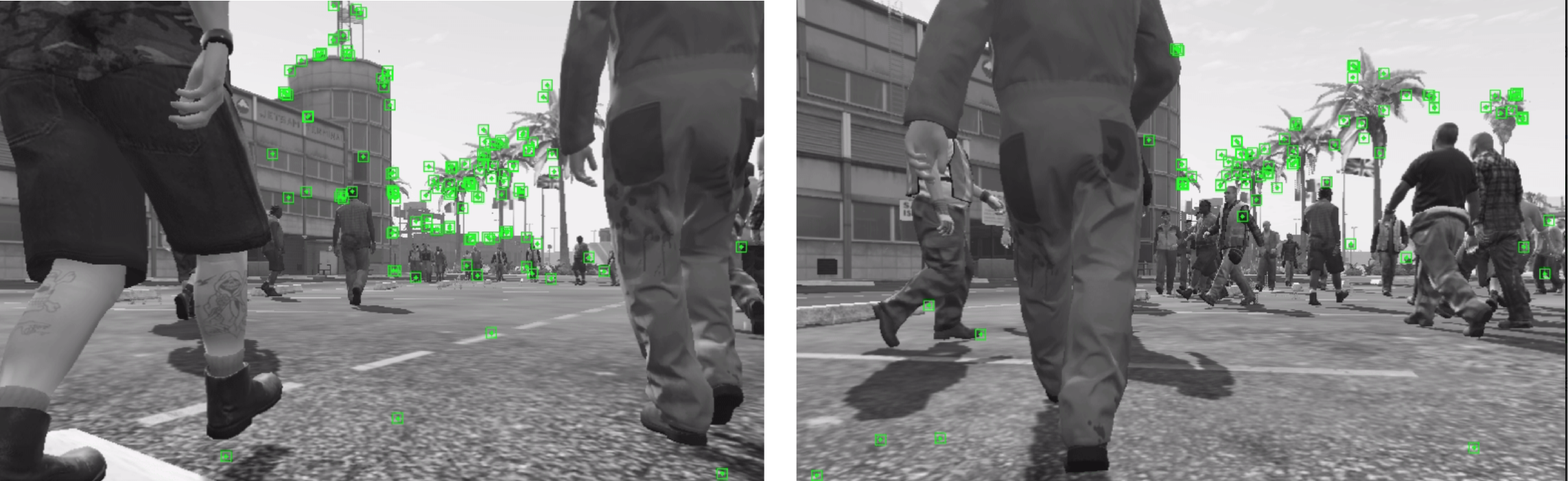}
    \caption{\textbf{Static keypoints trackable in crowded scenes.} In these typical two cases, static backgrounds far from the crowd are detected and tracked, while keypoints near the camera are untrackable due to severe occlusions. These static keypoints are at 30[m] from the observation camera.}
    \label{fig:farpoint}
\end{figure}

\section{Discussion}
In this paper, we propose a novel method for on-ground trajectory reconstruction of both the camera and pedestrians only from perceived movements of dynamic objects, \ie, pedestrians.
This allows us to recover the camera ego-motion even in a dense crowd, where static keypoints are occluded by surrounding pedestrians.
One may think ``even if the static keypoints near the camera are occluded and unable to track, we can still track backgrounds far from the crowd, \ie, buildings''.
\FIGREF{fig:farpoint} shows typical example cases in the GTAV dataset. The static backgrounds (\ie, buildings and trees) are detected and tracked, but are located far away ($\geq 30$ [m]) from the observation camera.

We simulate the robustness of a geometric relative pose solver~\cite{nister2004efficient} against noise according to the distances between the keypoints and the observation camera, and compare the accuracy with our approach based on dynamic keypoints.
We generate 100 static keypoints with uniform distribution in a voxel grid $\bm V_{w,h,d}$, where width $w=20$ [m], height $h=10$ [m], and depth $d=5$ [m].
The keypoints are captured by two cameras located at $[2:40]$ [m], where these two camera poses are randomly generated with conditions $\Delta r \leq  0.20$ [rad] and $\Delta \bm t \leq 1.0$ [m].
To test the robustness against detection noises in the image, we add uniform random noise ranging between [-1:1] [px] at each pixel and apply the five-point relative pose solver~\cite{nister2004efficient} with RANSAC~\cite{fischler1981random}.
At every distance from the camera, we test $100$ trials of relative pose solver for static keypoints randomly generated at each trial.

\FIGREF{fig:static-vs-dynamic} shows the errors of relative pose estimation for rotation $\Delta r$ and translation $\Delta \bm t$.
These results clearly show that the further the keypoints, the worse the accuracy of pose estimation. Although the relative pose solver performs well when static keypoints are observed at nearly $\leq 2$ [m], the translation errors are as worse as $0.50$ [m] on average when at $\geq 6$ [m].
We also compare these results with our view birdification results including detection noises by the external multi-object tracker~\cite{wang2019towards}. The red dotted line indicates the accuracy of our view birdification with multi-object tracking noises on the GTAV dataset (Table~\ref{tab:quant-gtav}).
Even with the detection noises, the errors of rotation and translation are $0.016$ [rad] and $0.097$ [m], respectively, which are significantly lower than those obtained with a relative pose solver with static keypoints observed from far away ($\geq 6$ [m]).

These results clearly indicate that \emph{nearby dynamic keypoints are better than distant static keypoints} for camera pose estimation in densely crowded environments. More specifically, if observed static keypoints are at as far as $\geq 6$ [m], our proposed view birdification based on the movements of nearby pedestrians performs better.

\begin{figure*}[t]
    \centering
    \includegraphics[width=0.9\linewidth]{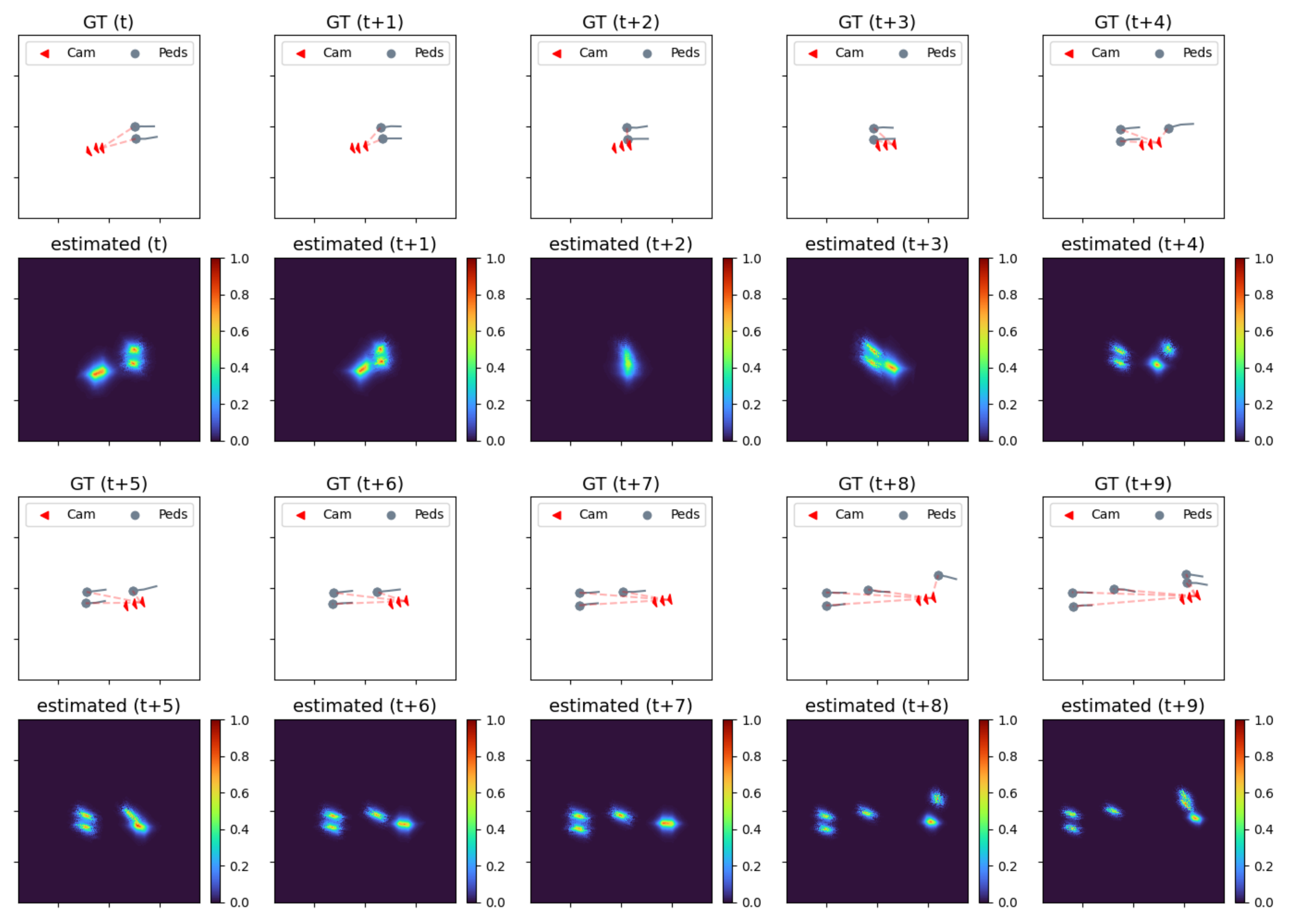}
    \caption{\textbf{Visualization of posterior distributions of the ETH dataset.} (First and third rows) Ground truth trajectories of the camera and its surrounding pedestrians. (Second and fourth rows) Visualization of posterior distributions of the location of the observer $\ped_0^\tau$ and surrounding pedestrians $\ped_k^\tau$. The heatmaps correspond to low (blue) to high (red) probabilities.}
    \label{fig:failure_case}
\end{figure*}

\section{Failure Cases}
We also analyze failure cases of our view birdification to understand the limitations of the method. For this, we picked sequences from \textbf{ETH} data that showed a high error rate in terms of camera localization.
\FIGREF{fig:failure_case} visualizes posterior distributions of the observer location
$p(\ped_0^\tau | \mathcal Z_{1:K}^\tau, \traj_{0:K}^{\tau-1})$
and surrounding pedestrians\\
$\int_{\ped_0^\tau \in \traj_s}
p(\traj_{1:K}^\tau | \mathcal Z_{1:K}^\tau, \ped_0^\tau) p(\ped_0^\tau)d\ped_0^\tau$
by sampling $\ped_0^\tau \in \traj_s$ in 
\EQREF{eq:height-gaussian} and \EQREF{eq:optimal-xk}, respectively.
The first and third rows depict the ground truth trajectories of the camera and pedestrians from $\tau$ to $\tau+9$.
The number of pedestrians changes from $K=3$ to $K=5$.
The second and fourth rows visualize the posterior distributions for each of those two rows.
As can be observed in the posteriors shown in the second row, the estimated observer location becomes a heavy-tailed distribution when the number of pedestrians in the crowd is small ($K=3$). In contrast, as shown in the fourth row, the posterior distribution becomes sharper when the crowd is denser ($K=5$).
The ambiguity of localization increases when pedestrians walk almost parallel to the observer (\eg, timesteps $\tau=\tau+2$ and $\tau+3$). In contrast, the posterior distribution becomes sharp again when the camera observes more pedestrians walking in diverse directions.

\section{Conclusion}
In this paper, we introduced view birdification, the problem of recovering the movement of surrounding people on the ground plane from a single ego-centric video captured in a dynamic cluttered scene. We formulated view birdification as a geometric reconstruction problem and derived a cascaded optimization approach that consists of camera ego-motion estimation and pedestrian localization while fully modeling the local pedestrian interactions. Our extensive evaluation demonstrates the effectiveness of our proposed view birdification method for crowds of varying densities.
Currently, occlusion handling is carried out by an external multi-object tracker. We envision a feedback loop from our birdificaiton framework that can inform the multi-object tracker to reason better about the occluded targets, which will likely enhance the accuracy as a whole even in heavily occluded scenes. 
We believe our work has implications for both computer vision and robotics, including crowd behavior analysis, self-localization, and situational awareness, and opens new avenues of application including dynamic surveillance.


\bibliographystyle{spmpsci}      
\bibliography{main} 

\end{document}